\documentclass{article}
\usepackage[preprint]{neurips_2025}
\usepackage[utf8]{inputenc} 
\usepackage[T1]{fontenc}    
\usepackage{hyperref}       
\usepackage{url}            
\usepackage{booktabs}       
\usepackage{amsfonts}       
\usepackage{nicefrac}       
\usepackage{microtype}      
\usepackage{xcolor}         
\usepackage{amsmath}
\usepackage{amssymb}
\usepackage{mathtools}
\usepackage{amsthm}

\usepackage[capitalize,noabbrev]{cleveref}
\theoremstyle{plain}

\theoremstyle{definition}

\theoremstyle{remark}

\usepackage[textsize=tiny]{todonotes}
\usepackage{microtype}
\usepackage{graphicx}
\usepackage{subfigure}
\usepackage{booktabs} 
\usepackage{subcaption}
\usepackage{hyperref}
\usepackage{colortbl}
\usepackage[export]{adjustbox}
\usepackage{wrapfig}
\usepackage[dvipsnames]{xcolor}
\usepackage[most]{tcolorbox}
\usepackage{caption}
\usepackage{pifont}
\usepackage{mathrsfs}
\usepackage{colortbl}
\definecolor{mygray}{gray}{.9}
\usepackage{makecell}
\usepackage{multirow}
\usepackage{enumitem}

\title{Can Data-Driven Dynamics Reveal Hidden Physics? \\
There Is A Need for Interpretable Neural Operators}

\author{
Wenhan~Gao\textsuperscript{1}\thanks{Equal contribution.}, \hspace{0.35em}
Jian~Luo\textsuperscript{1, 2}\footnotemark[1], \hspace{0.35em}
Fang~Wan\textsuperscript{2}, \hspace{0.3em}
Ruichen~Xu\textsuperscript{1}, \\
\textbf{Xiang~Liu\textsuperscript{2}, \hspace{0.3em}
Haipeng~Xing\textsuperscript{1}, \hspace{0.3em}
Yi~Liu\textsuperscript{1, 2}} \\
\textsuperscript{1}Department of Applied Mathematics and Statistics, Stony Brook University \\
\textsuperscript{2}Department of Computer Science, Stony Brook University \\
Correspondence to: \texttt{yi.liu.4@stonybrook.edu}
}

\begin{document}

\maketitle
\begin{abstract}
Recently, neural operators have emerged as powerful tools for learning mappings between function spaces, enabling data-driven simulations of complex dynamics. Despite their successes, a deeper understanding of their learning mechanisms remains underexplored. In this work, we classify neural operators into two types: (1) Spatial domain models that learn on grids and (2) Functional domain models that learn with function bases. We present several viewpoints based on this classification and focus on learning data-driven dynamics adhering to physical principles. Specifically, we provide a way to explain the prediction-making process of neural operators and show that \textbf{neural operator can learn hidden physical patterns from data}. However, this explanation method is limited to specific situations, highlighting the \textbf{urgent need for generalizable explanation methods}. Next, we show that a simple dual-space multi-scale model can achieve SOTA performance and \textbf{we believe that dual-space multi-spatio-scale models hold significant potential to learn complex physics and require further investigation.} Lastly, we discuss \textbf{the critical need for principled frameworks to incorporate known physics into neural operators}, enabling better generalization and uncovering more hidden physical phenomena. 
\end{abstract}

\begin{figure*}[h!]
    \centering
    \subfigure[]{\includegraphics[width=0.47\textwidth]{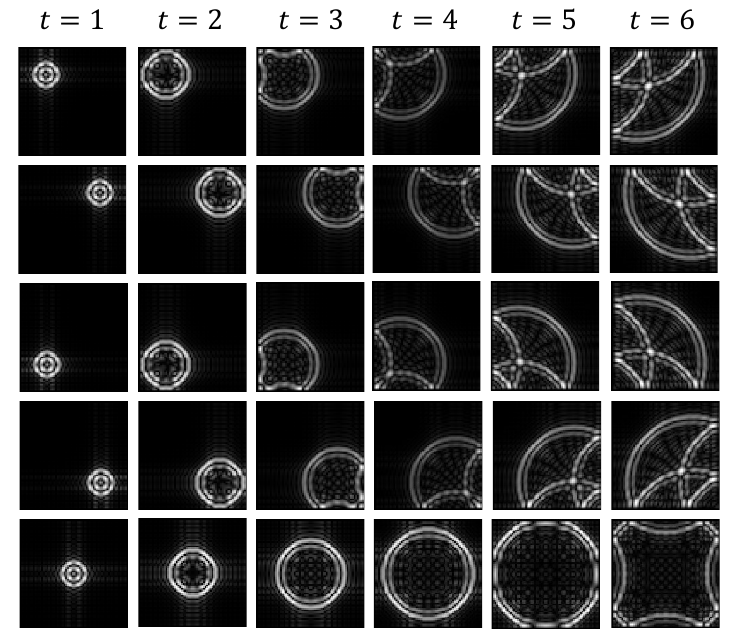}} 
    \subfigure[]{\includegraphics[width=0.47\textwidth]{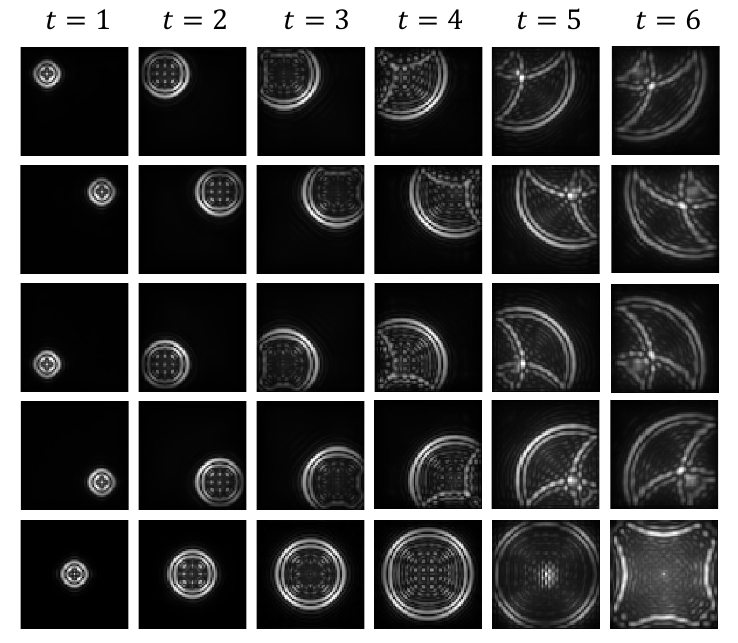}} 
    \vspace{-10pt}
    \caption{Comparison of spatial dependencies of a wave propagation at different times: (a) Analytical spatial dependencies derived from the governing equations. (b) Spatial dependencies learned by a well-trained neural operator, demonstrating its ability to approximate the underlying spatial relationships. Spatial dependencies, defined in Appendix \ref{append:spatial_dependencies_and_wave_analytical}, describe which locations in the input domain have the greatest influence on the output, serving as a fundamental characteristic of many physical systems. For wave propagation, they reveal how disturbances at various input locations influence the resulting wave behavior, with isotropic patterns under certain conditions.} \label{fig: header_learned_and_analytical_spatial}
    \vspace{-18pt}
\end{figure*}

\section{Introduction}
\vspace{-5pt}
Researchers in various fields of science and engineering seek to explore the behavior of physical systems under varying parameters, such as different initial conditions, forcing sources, or coefficient functions. Simulating these systems is demanding due to the prohibitive computational cost associated with numerical solvers. Moreover, for many engineering problems, the exact governing equations are difficult to derive~\citep{ShuiwangJi_AI4SCI_REVIEW}. Recent advances in deep learning motivate the use of neural networks to learn data-driven dynamics; these methods are loosely referred to as neural operators (NOs). Unlike numerical solvers that solve one instance at a time, neural operators learn the mapping from the input function space to the solution function space. After the training phase, performing inference only necessitates a forward pass of the network, which can be several orders of magnitude faster than traditional numerical methods. Over the past few years, various neural operator architectures have been proposed and shown great performance, such as the Fourier Neural Operator (FNO)~\citep{FNO}, Convolutional Neural Operator (CNO)~\citep{convolutional_NO}, Deep Operator Network (DeepONet)~\citep{DeepOnet}, and many transformer-based neural operators~\citep{Galerkin_transformer, FactFormer, transolver}. As data-driven methods, neural operators have been of great use in many fields where exact governing equations may or may not be known, including climate modeling~\citep{pathak2022fourcastnet} and fluid dynamics~\citep{fluid_survey}.

While neural operators have achieved great successes in practice and are driving significant real-world impacts, their learning mechanisms are underexplored. There have been several efforts in this line of research, for example, function space optimization for scientific machine learning tasks~\citep{optimization_for_sciML}, universal approximation~\citep{kovachki2021universal, DeepOnet}, and discretization errors~\citep{discretization_error_FNO, gao2025discretizationinvariance}. However, fundamental questions about the interpretability of neural operators, such as whether they can learn the underlying physics, persist.
Neural operators can be categorized into spatial domain models and functional domain models, where spatial domain models operate directly on the spatial space with grid-based representations (e.g., CNO~\citep{convolutional_NO} and Transformer-based NOs~\citep{Galerkin_transformer}), while functional domain models focus on learning with functional bases (e.g.,  DeepONet~\citep{DeepOnet} and T1~\citep{T1_NO}). Some operators can be a hybrid of both; for example, FNO~\citep{FNO} operates as a grid-based method while its learned parameters predominantly are in the spectral domain. Different neural operators possess different properties and may excel in different tasks depending on their design. However, there is currently a lack of principled research on how neural operators learn (explainable neural operator) and to what extent neural operators can learn the underlying physics hidden in the data. In Fig. \ref{fig: header_learned_and_analytical_spatial}, we present an example of CNO learning the wave propagation process. It is obvious that neural operators can effectively capture the spatial dependencies of the wave propagation process.
 
A deeper understanding of how neural operators learn could, in turn, guide the development of more effective architectures better suited to scientific applications. Specifically, we highlight a few empirical observations, propose conjectures based on these observations, and reinstate our position. \textit{For each viewpoint, we summarize our sub-positions, as well as observations and conjectures if applicable, in colored text-boxes at the end.} We present our current findings and position, but we are certain there is much more to uncover and much more effort should be invested.

\begin{tcolorbox}[colback=white, before skip=0.2cm, after skip=0.2cm, boxsep=0.0cm, middle=0.1cm, top=0.1cm, bottom=0.1cm]
\textbf{\textit{Overall Position.}} Neural operators can learn hidden physics from data, such as the spatial dependencies. However, their effectiveness depends heavily on the design of the architecture. Therefore, there is a pressing need for interpretable neural operators, which includes two aspects: \ding{182} Explanation methods that study how neural operators make decisions. \ding{183} Better neural operator designs that can more effectively learn physics and further improve interpretability.
\end{tcolorbox}

\section{Background}
In this section, we provide an overview of operator learning in Sec.~\ref{sec:operator_learning}  and introduce several architectures that are discussed in this work, in Sec.~\ref{sec:architectures}.

\subsection{Operator Learning}\label{sec:operator_learning}
Recent advancements in deep learning have motivated the use of neural networks to learn operators (mappings between function spaces)~\citep{NO_survey_JMLR}. Neural operators play a crucial role in scientific computing, where they are applied to simulate complex physical phenomena that otherwise require extensive numerical computations. A particular use of neural operators is to approximate the solution operator of parametric PDEs. Consider a parametric PDE of the form:
\begin{equation}\label{eq: general_pde}
\begin{aligned}
\mathcal{N}(a, u)(x)&=f(x), & & x \in \Omega \\
u(x) & =0, & & x \in \partial \Omega
\end{aligned}
\end{equation}
where $\Omega \subset \mathbb{R}^d$ is the domain, a bounded open set, $\mathcal{N}$ is a differential operator that is possibly non-linear, $a \in \mathcal{A}$ is the parametric input function, $f$ is a given fixed function in an appropriate function space determined by $\mathcal{N}$, and $u \in \mathcal{U}$ is the solution function. The PDE solution operator is defined as $G(a)=u$. 

For appropriate function spaces $\mathcal{A}$ and $\mathcal{U}$, we are interested in learning an operator $G: \mathcal{A} \mapsto \mathcal{U}$ with a neural operator $\mathcal{G}(\cdot; \theta)$ with $\theta \in \Theta$ being the network parameters through a finite collection of observations $\left\{a_j, u_j\right\}_{j=1}^{N_{\text{data}}}$ with $a_j \sim \mu$ being i.i.d. samples drawn from some probability measure $\mu$ supported on $\mathcal{A}$. We aim to control the $L_\mu^2(\mathcal{A} ; \mathcal{U})$ Bochner norm of the approximation on average with respect to $\mu$ as the learning objective of the neural operator:
\begin{equation}
\begin{aligned}
 \min _{\theta \in \Theta} \|G(\cdot)-\mathcal{G}(\cdot ; \theta)\|_{L_\mu^2(\mathcal{A} ; \mathcal{U})}^2 &=\mathbb{E}_{a \sim \mu}\|G(a)-\mathcal{G}(a ; \theta)\|_{\mathcal{U}}^2\\
&=\int_{\mathcal{A}}\|G(a)-\mathcal{G}(a ; \theta)\|_{\mathcal{U}}^2 d \mu(a) \\
& \approx \frac{1}{N_{\text{data}}} \sum_{j=1}^{N_{\text{data}}}\left\|u_j-\mathcal{G}_\theta\left(a_j\right)\right\|_{\mathcal{U}}^2.
\end{aligned}
\end{equation}

Practically, discretized forms of these functions are given, and we seek to approximately solve the empirical-risk minimization problem directly at those discretization points, typically using $\ell_1$ or $\ell_2$ errors \citep{convolutional_NO, FNO} as evaluation metrics. In this paper, we generally refer to any network architecture employed for operator learning tasks as neural operators. 


\subsection{Architectures} \label{sec:architectures}
\textbf{FNO (Hybrid). } Fourier neural operator (FNO)~\citep{FNO} is a special integral neural operator. Inspired by linear layers in feed-forward neural networks, each integral neural operator layer consists of a fixed non-linearity and a linear integral operator $\mathcal{K}$ modeled by network parameters, defined as $(\mathcal{K} v)(x)=\int \kappa_\theta(x, y) v(y) \mathrm{d} y$. FNO further imposes translation invariance on the kernel, $\kappa_\theta(x, y)=\kappa_\theta(x-y)$, resulting in a convolution operator. Convolution can be efficiently carried out as element-wise multiplication in the frequency domain:
\vspace{-3pt}
\begin{equation}\label{fouerier_conv}
(\mathcal{K} v)(x)=\mathcal{F}^{-1}(\mathcal{F} \kappa_\theta \cdot \mathcal{F} v)(x),
\end{equation}
where $\mathcal{F}$ and $\mathcal{F}^{-1}$ are the Fourier transform and its inverse, respectively. FNO directly learns complex weights $\mathcal{F} \kappa_\theta$ in the frequency domain. FNO operates on functional domains to learn global continuous kernel integrals while incorporating non-linearities directly in the spatial domain. 

\textbf{DeepONet (Functional). } Deep operator network (DeepONet)~\citep{DeepOnet} is based on the universal approximation theorem for operators~\citep{Chen1995Universal_approximation_operator}. DeepONet comprises two neural networks: a \textit{branch network} that processes input functions $v$ to produce function coefficients, $b_i$, and a \textit{trunk network} that processes evaluation points $x$ and produces values of the learned basis functions at these points, $t_i(x)$. The output of DeepONet is computed as the dot product of their outputs, $G(v)(x) = \sum_{i=1}^p b_i \cdot t_i(x) $, where $p$ is the number of basis functions. 

\textbf{T1 (Functional).} Transform Once (T1)~\citep{T1_NO} is an efficient frequency-domain learning model that avoids the high cost of repeated Fourier transforms in FNO by using a single Fourier or cosine transform and performing all learning directly in the frequency domain. 

\textbf{CNO (Spatial). } Convolutional neural operator (CNO)~\citep{convolutional_NO, ReNOs} extends U-Net by adding additional up-and-down sampling in the activation functions to mitigate aliasing errors. Given a learning task between band-limited function spaces, the activation function introduces arbitrarily high frequencies, breaking the band limit. Up-and-down sampling in CNOs first lifts the function to a higher band-limit by up-sampling and, after the activation function is applied, projects the function back to the original band-limited space by down-sampling.

\textbf{Transformer NO (Spatial). } Transformer-based architectures~\citep{Galerkin_transformer, DPO_transformer, GNO_transformer, transolver, nonlocal_attention_operator} are inspired by natural language processing tasks, where the inputs have varying lengths. This aligns with the resolution invariance property desired in neural operators. In the context of operator learning, the self-attention mechanism can be adapted as a learned integral operator. In particular, we will use Galerkin Transformer (GT-former)~\citep{Galerkin_transformer} as an example in this work. More details are given in Appendix \ref{append: attention_no}.

\section{Uncover Learned Physics: A Spatial Dependency Perspective}
\begin{wrapfigure}[29]{r}{0.48\textwidth}
    \centering
    \vspace{-15pt}
    \includegraphics[width=0.99\linewidth]{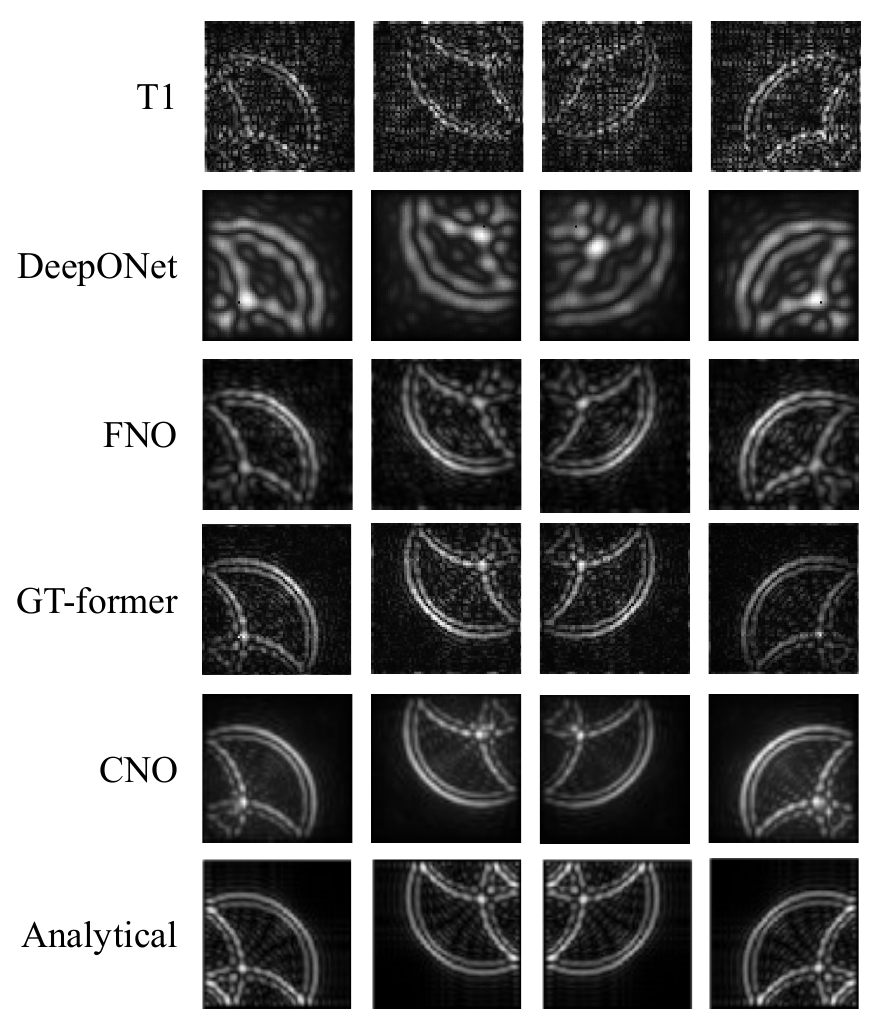}
    \caption{Comparison of the analytical ERF function and that of the learned neural operators at various locations. The spatial domain models, CNO and GT-former, learn the wave patterns well. In contrast, the functional domain models, T1 and DeepONet, struggle to capture wave patterns. The hybrid model, FNO, is between these two models; it learns fairly well but introduces noise outside the arc region.} \label{fig:wave_cno_vs_fno}
\end{wrapfigure}

To examine whether neural operators learn data-driven dynamics in a way that adheres to the underlying physics, we analyze the spatial dependencies captured as an example to explain how neural operators make decisions. Spatial dependencies, which encode how physical quantities interact across space, serve as a fundamental characteristic of many physical systems, such as electromagnetic field interactions, heat diffusion, and fluid dynamics \citep{PDE_evans10}. To study the learned spatial dependencies from neural operators, we adopt the notion of the effective receptive field (ERF) from \citet{Effective_receptive_field} for operators $G: a \mapsto u$.

For simplicity, we assume one-dimensional (a single channel) input functions $a(x): D \mapsto \mathbb{R}$ and output functions $u(x'): D \mapsto \mathbb{R}$ of the same domain $D$. The ERF function $\operatorname{erf}: D \times D \mapsto \mathbb{R}$ is defined as the functional derivative: $\operatorname{erf}(x', x) = \frac{\partial u}{\partial a}(x', x)$. For a specific output location $x_o \in D$, we have $\operatorname{erf}_{x_0}(x) = \frac{\partial u}{\partial a}(x', x)|_{x' = x_0}$, which describes which locations in the input domain have the greatest influence on the output. More details can be found in Appendix \ref{append:spatial_dependencies_and_wave_analytical}. We conduct experiments to study spatial dependencies through ERFs on five representative architectures, T1, DeepONet, FNO, GT-former, and CNO, which span function domain models (T1, DeepONet), spatial domain models (GT-former, CNO), and a mix of both (FNO). We focus on the wave and Navier-Stokes equations for this analysis, as they are representative PDEs with rich and contrasting spatial dependency structures. Wave exhibits long-range interactions due to wave propagation over time, while NS can involve both localized and global interactions governed by advection and diffusion.

\subsection{Wave Propagation} \label{sec:wave}
We consider the wave equation from \citep{convolutional_NO} given by
\begin{equation}
\begin{aligned}
    & u_{tt} - c^2 \Delta u = 0, && \text{in } D \times (0, T), 
    \\
    & u_0(x, y) = f(x, y), && \text{for } (x, y) \in D
\end{aligned}
\end{equation}
with a constant propagation speed $c=0.1$. The underlying operator $\mathcal{G}_t: f \mapsto u( \cdot, t)$ maps the initial condition $f$ into the solution at a later time $t$. More details on this equation, data generation, and the experimental details can be found in Appendix \ref{append: Wave}. In terms of wave propagation, the ERF function indicates how disturbances at different input locations affect the wave behavior. We can obtain the ERF function analytically for this specific wave equation example in the following form:
\begin{equation}
\begin{aligned}
    \frac{\partial u}{\partial u_0}(x',y',x,y,t)=\lim_{K\rightarrow \infty}\sum^{K}_{i,j=1} \Big[\sin(\pi i x') \sin(\pi j y')  \sin(\pi ix) \sin(\pi jy) \cos(c\pi t\sqrt{i^2+j^2})\Big].
\end{aligned}
\end{equation}
The derivation is provided in Appendix \ref{append:analytical_ERF}. We perform experiments to study the learned spatial dependencies. All models are well-trained and achieve a reasonable $\ell_2$ error. The visual results are presented in Fig. \ref{fig:wave_cno_vs_fno} with quantitative results in Appendix \ref{append: Wave}. 

Clearly, CNO and GT-former, as spatial-domain models, demonstrate superior capability in capturing wave patterns compared to all other models. The hybrid model FNO retains all intermediate representations on a spatial grid and applies activations in the spatial domain; its Fourier components primarily serve to accelerate spatial convolutions. As a result, FNO maintains strong spatial learning capabilities and clearly outperforms the two functional domain models in capturing wave patterns. DeepONet and DeepOnet, as functional domain models, fail to faithfully capture these spatial dependencies, i.e., the dynamics of arced wave patterns. \textbf{These results confirm that models grounded in spatial representations can better capture wave patterns.}

\textit{Additionally, it should be noted that standard evaluation metrics, like $\ell_p$ loss, do not necessarily reflect whether the learned neural operators adhere to the underlying dynamics.} As shown in Fig. \ref{fig:wave_cno_vs_fno}, while FNO achieves a better $\ell_2$ error than GT-former ($2.77\%$ compared to $3.00\%$), FNO is noticeably worse at capturing wave patterns, as there appear to be many noisy bright dots across the entire domain. We provide another example of this phenomenon in Appendix~\ref{append: Wave}.

\textbf{\begin{tcolorbox}[before skip=0.2cm, after skip=0.2cm, boxsep=0.0cm, middle=0.1cm, top=0.1cm, bottom=0.1cm]
\textit{\textbf{Observation.} The commonly used $\ell_2$ error does not necessarily indicate how well the prediction-making process of neural operators conforms to the underlying physics.}
\end{tcolorbox}}

\subsection{The Navier-Stokes Equation} \label{sec:ns}
To further investigate the ability of neural operators in learning spatial dependencies, we provide the ERF results on the incompressible Navier-Stokes equation from \citet{FNO}. More details on this equation, data generation, and experiments can be found in Appendix \ref{append: NS_equation}. We are interested in learning the operator $G_t: w_0 \mapsto w_t$ mapping the initial vorticity to the vorticity at a later time $t$. Unlike the wave equation, where the analytical ERF function can be derived, we can only provide the empirical ERF function for the learned neural operators. 

Although the ERFs of the underlying operator lack an analytical form, we perform experiments under two specific settings where the results can be meaningfully interpreted:
\begin{itemize}
    \item Varying Time Steps: We are interested in learning the operators $G_{5}, G_{10}$ and $G_{15}$ respectively. As the time scale gets longer, we expect to capture a larger effective receptive field as the disturbance has more time to propagate through the domain.
    \item Varying Reynold Number (Vorticity Constant): As the Reynolds number increases (vorticity constant decreases), the flow exhibits finer structures (small eddies) and sharper gradients. The dependencies between spatial locations become more localized.
\end{itemize}

The results of CNO (spatial) and T1 (functional) are shown in Fig. \ref{fig:NS_cno_vs_fno}. Additional results on other neural operators are shown in Appendix \ref{append: NS_equation}. While an analytical solution is not available, the spatial dependencies captured by spatial domain models appear reasonable and align with the preceding discussion. In contrast, functional domain models produce results that are clearly unrealistic and fail to capture the underlying spatial dynamics. In addition, the hybrid model FNO also produces physically plausible results, potentially because its hybrid nature allows certain patterns to be better captured in the spectral domain. However, T1 operates entirely in the spectral domain and still fails to capture the dynamics, suggesting that the performance of FNO primarily benefits from its spatial characteristics. Moreover, it should be noted that the dataset is generated using spectral methods, which could potentially introduce errors and biases in the spectral representations that favor FNO. As there is no analytical form, the underlying mechanisms remain unclear. 
\begin{figure*}[h]
\vspace{-8pt}
    \centering
    \includegraphics[width=0.9\linewidth]{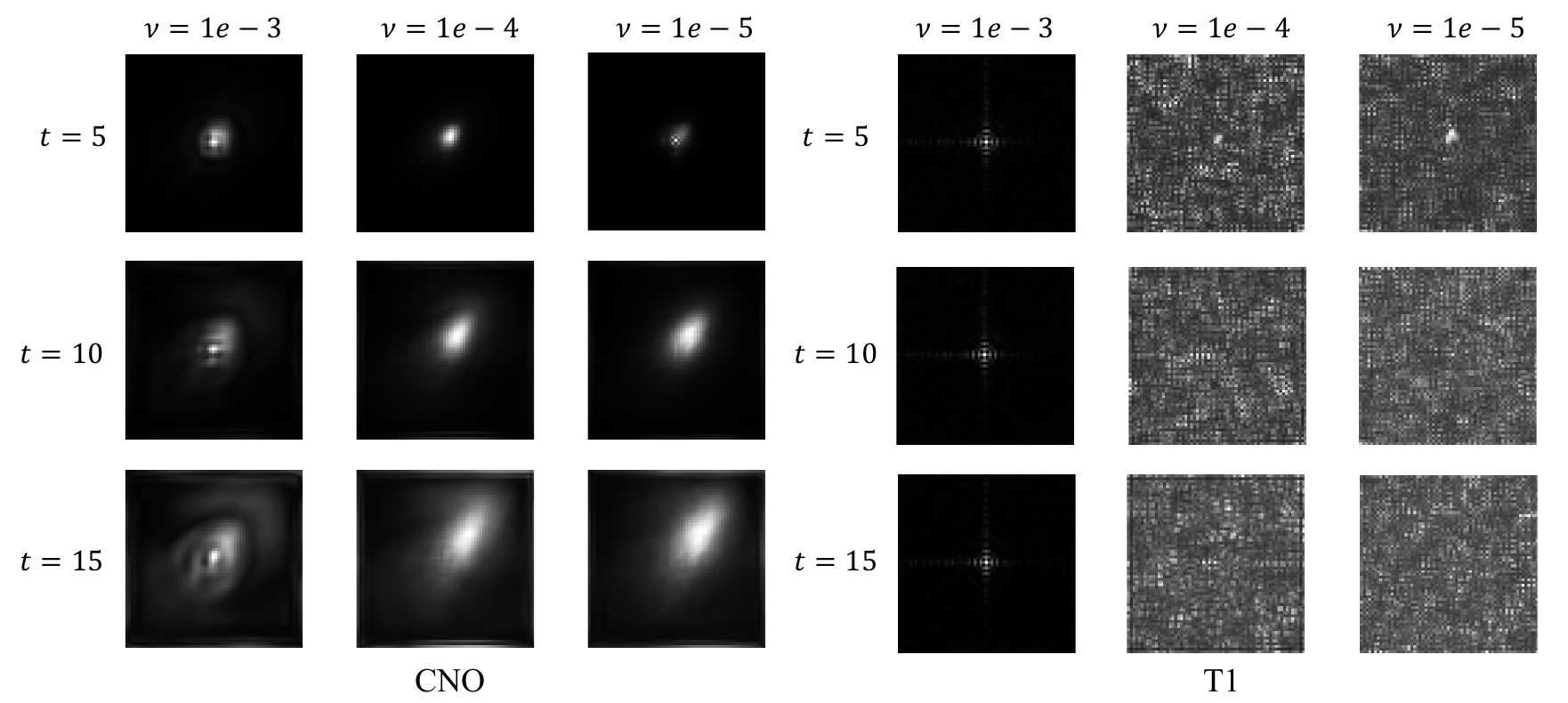}
    \vspace{-8pt}
    \caption{Learned spatial dependencies on the incompressible Navier-Stokes equation reveal insightful patterns. The learned spatial dependencies for all spatial domain models (see Appendix \ref{append: NS_equation} for more results) align with expectations: as the vorticity constant $\nu$ decreases, local patterns become more significant, whereas global patterns gain importance with prolonged time scales.} \label{fig:NS_cno_vs_fno}
    \vspace{-21pt}
\end{figure*}

\textbf{Clearly, for both Wave and Navier-Stokes equations, spatial domain models better capture spatial dependencies.} This phenomenon highlights a key limitation of functional domain models in scenarios where spatial interactions play a significant role. \textit{We hypothesize that the choice of the learning domain and the function bases for functional domain models plays a crucial role in capturing the underlying physics. Certain physical phenomena may be easier to learn or more evident in the spatial domain or with specific functional bases.}

\begin{tcolorbox}[before skip=0.2cm, after skip=0.2cm, boxsep=0.0cm, middle=0.1cm, top=0.1cm, bottom=0.1cm]
\textit{\textbf{Observation.} Spatial domain models (e.g. CNO and GT-former) effectively learn the spatial dependencies, whereas functional domain models (e.g. DeepONet and T1) struggle to fully capture the intrinsic spatial patterns. The hybrid model, FNO, captures overall spatial dependencies well but may exhibit local inaccuracies or noises.}
\end{tcolorbox}

\begin{tcolorbox}[colback=pink, before skip=0.2cm, after skip=0.2cm, boxsep=0.0cm, middle=0.1cm, top=0.1cm, bottom=0.1cm]
\textit{\textbf{Conjecture.} Spatial domain models are inherently more suited to capturing spatial dependencies. In contrast, functional domain models may be more effective in learning physical principles that are more naturally expressed or transparent in certain function spaces.}
\end{tcolorbox}

\begin{tcolorbox}[colback=Apricot, before skip=0.2cm, after skip=0.2cm, boxsep=0.0cm, middle=0.1cm, top=0.1cm, bottom=0.1cm]
\textit{\textbf{Sub-position.} We should investigate the fundamental differences in learning mechanisms between spatial and functional domain models, as well as among various bases within functional domain models.}
\end{tcolorbox}

It is important to note that, in most cases, the analytical ERF functions are intractable to compute, or at times, the PDE forms are not even provided. Moreover, the numerical ERF computation through Autograd only works for explaining spatial relationships; in some operator learning applications, spatial dependencies may not play a critical role or even be applicable. \textit{In summary, there are two major limitations of the ERF-function-based explanation: \ding{182} The analytical forms of the ERF function are intractable in most cases; \ding{183} Spatial dependencies cannot serve as an ``explainer'' in many applications when spatial relationships are far less significant. These limitations underscore the need for more general explanation methods.}

\begin{tcolorbox}[colback=Apricot, before skip=0.2cm, after skip=0.2cm, boxsep=0.0cm, middle=0.1cm, top=0.1cm, bottom=0.1cm]
\textit{\textbf{Sub-position.} Developing effective explanation methods is essential for gaining deeper insights into how neural operators capture the underlying physics. There is a pressing need for generalizable explanation methods for operator learning that are independent of the governing equations and can be applied to more operator architectures.}
\end{tcolorbox}
\vspace{-3pt}
\section{Uncover the Learning Process: A Multi-spatio-scale Perspective} \label{sec: kernel}
\vspace{-3pt}

Many operator learning tasks involve capturing complex multi-scale features, making it crucial for neural operators to effectively model both local and global patterns~\citep{Multi-spatiotemporal-scale_PDE}. Local convolution kernels that are usually employed in computer vision tasks are restricted to small spatial regions. Most operator learning applications exhibit global spatial dependencies, making local convolution kernels insufficient for learning. One potential mitigation for this issue is to use U-Net based architectures, which leverage local kernels to capture global features effectively through their down-sampling and up-sampling paths. The down-sampling path captures hierarchical features and encodes global context by progressively reducing the spatial resolution, while the up-sampling path restores the resolution and integrates these global features with local details through skip connections. \textbf{Thus, spatial domain models with local kernels, e.g. CNO, can learn multi-spatio-scale features in similar manners. However, the down-sampling paths involve an information compression process that may lead to a loss of fine-grained details.} This compression can be particularly limiting when modeling phenomena that require precise localities.

Functional domain models, on the other hand, naturally process both local and global information with suitable function bases. 
Taking T1 as an example, global and local features are captured through low and high frequency modes, respectively. \textbf{In practice, most functional domain models, even including the hybrid model FNO, perform well when capturing smooth, global structures but often struggle to accurately represent sharp variations or localized fine-scale features.} In~\citet{localized_kernel_FNO}, the authors argue that global operations (e.g. in FNO, a single Fourier mode is global in the spatial domain) may fail to capture local details. They propose mitigating this issue by incorporating localized kernels that converge to the differential operator as the resolution approaches infinity. Moreover, as shown in~\citet{Better_understanding_spectral_perspective}, FNO still, surprisingly, shows low-frequency bias to a large extent even without truncating the high frequency modes.

Building on these claims, \textit{we discover that the advantages of integrating global Fourier filters with local convolutions extend beyond merely capturing high-frequency local details or addressing spectral bias.} To validate this, we conducted experiments on the Darcy flow equation dataset from~\citet{FNO}, the Helmholtz equation dataset from~\citet{helm1}, and the Wave and Allen-Cahn equations form~\citet{convolutional_NO}. Details about these datasets are provided in Appendix~\ref{append: exp}.

\begin{wraptable}[12]{r}{0.60\textwidth}
    \centering
    \small
    \vspace{-14pt}
    \caption{Results with the original FNO ($12$ modes), the FNO enhanced with local convolutions (FNO$_{3\times 3}$), and the FNO with full modes (FNO$_\text{full}$). The relative $\ell_2$ error and standard deviation are reported. The number of network parameters scales with the data resolution. \textbf{For reference, we report parameter counts in millions for the Darcy flow dataset.}}
    \vspace{-4pt}
    \begin{tabular}{lccc}
    \Xhline{2\arrayrulewidth}
    \textbf{Dataset} 
    & \makecell[c]{FNO\\(2.38M)} 
    & \makecell[c]{FNO$_{3\times 3}$\\(2.45M)} 
    & \makecell[c]{FNO$_\text{full}$\\(16.79M)} 
    \\ \hline
    Darcy       & $0.69(\pm 0.03)$     & $\mathbf{0.51(\pm0.01)}$     & $0.65(\pm0.02)$ \\
    Helmholtz   & $0.68(\pm 0.01)$        & $\mathbf{0.53(\pm 0.01)}$        & $0.66(\pm 0.02)$ \\
    Wave & $2.77(\pm 0.13)$     & $\mathbf{2.08(\pm0.06)}$     & $2.73(\pm0.12)$ \\
    Allen-Cahn  & $0.42(\pm 0.02)$     & $\mathbf{0.19(\pm0.01)}$     & $0.26(\pm0.02)$ \\
    \Xhline{2\arrayrulewidth}
    \end{tabular}
    \label{tab:res_multi_scale}
\end{wraptable}

To investigate the effect of combining local convolution filters and global Fourier filters, we add extra $3 \times 3$ convolutions in the residual connection path of FNO, which barely increases the number of parameters, e.g., from $2.38$ millions to $2.45$ millions.
Additionally, we study whether capturing high-frequency details is the key factor for improved performance by comparing it with the FNO model trained with all modes, which has, as a reference, $16.79$ million parameters for the Darcy flow equation The results are presented in Table~\ref{tab:res_multi_scale}. 

The results clearly show that simply adding local convolutions leads to significant improvements; the improved performance cannot be solely attributed to the ability to learn high-frequency local details or spectral bias as the FNO with full mode does not improve the performance by much. Moreover, we provide the spectral plot of the error distribution for the Darcy flow example in Fig.~\ref{fig:spectral_multiscale} in Appendix~\ref{append:darcy_data}; clearly, the inclusion of local convolution kernels not only reduces errors in the high-frequency modes but also in the low-frequency modes.
We suspect the reason lies in FNO's limited local learning ability. As a hybrid model with most parameters in the Fourier space, any change to its coefficients leads to global effects. While Fourier representations can encode both local and global patterns, the learning process remains predominantly global. In contrast, the dual-space multi-scale model (the FNO enhanced with local convolutions) likely learns hierarchically, capturing global trends in the functional domain before refining local details through spatial convolutions. However, the underlying reasons are not fully understood, and further research on dual-space multi-scale learning is needed.
 
\begin{tcolorbox}[before skip=0.2cm, after skip=0.2cm, boxsep=0.0cm, middle=0.1cm, top=0.1cm, bottom=0.1cm]
\textit{\textbf{Observation.} We observe a notable performance improvement by simply incorporating a few local convolution layers into the FNO architecture.}
\end{tcolorbox}

\begin{tcolorbox}[colback=pink, before skip=0.2cm, after skip=0.2cm, boxsep=0.0cm, middle=0.1cm, top=0.1cm, bottom=0.1cm]
\textit{\textbf{Conjecture.} The combination of global Fourier filters in the spectral domain, combined with local convolution filters in the spatial domain, enable multi-scale spatio-hierarchical representation learning, which plays a key role in the observed performance gains.}
\end{tcolorbox}

\begin{tcolorbox}[colback=Apricot, before skip=0.2cm, after skip=0.2cm, boxsep=0.0cm, middle=0.1cm, top=0.1cm, bottom=0.1cm]
\textit{\textbf{Sub-position.} We believe that integrating functional and spatial learning holds significant potential. More studies should be conducted on dual-space multi-scale learning, including both designing advanced architectures and explaining why and when it brings better performance. }
\end{tcolorbox}

\section{Incorporating Inductive Biases for Improved Interpretability}
As we emphasize the need for interpretable neural operators, another crucial perspective is the incorporation of known inductive biases (physics priors). Inductive biases enhance interpretability by constraining the model’s hypothesis space to a physically meaningful solution space and guiding the model to learn coupled or emergent physics. In Sec.~\ref{sec:PI_ML}, we introduce physics-informed losses, discuss their limitations, and advocate for inductive biases in network designs. In Sec.~\ref{sec:physics_by_design}, we discuss equivariant neural operators as a notable example of incorporating inductive biases into the model and improving the interpretability of the learned neural operator. These physics priors can be applied to either functional or spatial models, depending on the task, although certain priors are more naturally incorporated into either functional or spatial domains, as we discuss in Sec.~\ref{sec:physics_by_design}.

\begin{wrapfigure}[13]{r}{0.46\textwidth}
\vspace{-40pt}
    \centering
    \includegraphics[width=0.99\linewidth]{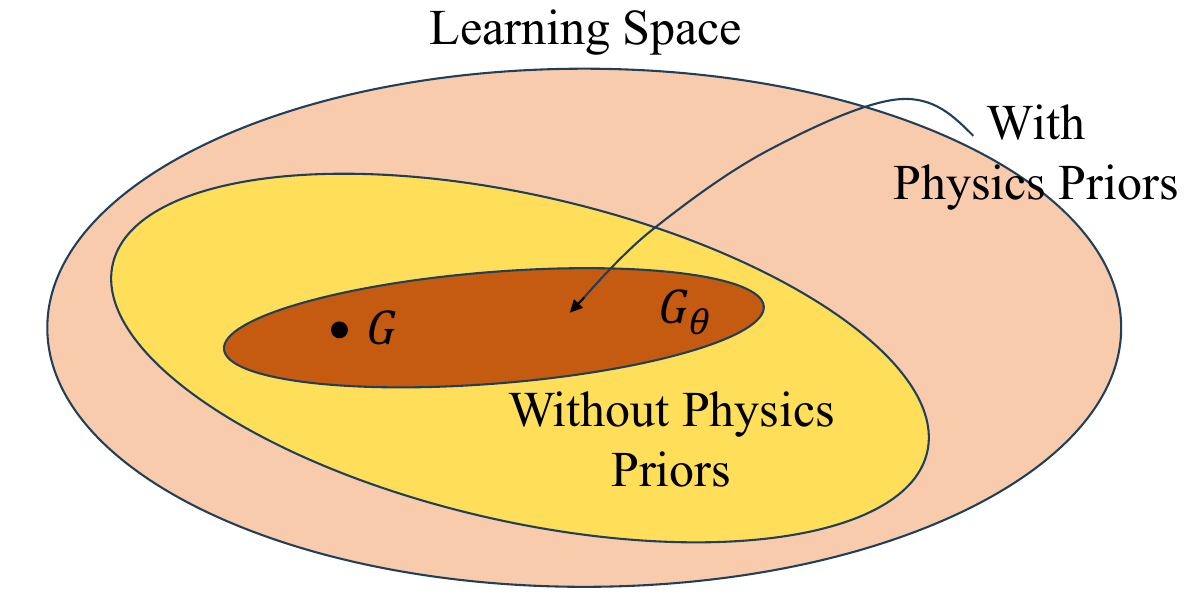}
    \caption{Enforcing physics priors into the network design constraints the learning space to be a subspace of operators that respect those physics priors. This reduces the need of learning data and enables better generalization, leading to more robust and interpretable results. } \label{fig:physical_prior_learning}
\end{wrapfigure}

\subsection{Physics-informed Constraints} \label{sec:PI_ML}
One promising approach of incorporating known physics is the incorporation of physics-informed constraints directly into the loss function as a supervising signal penalizing physical inconsistency \citep{PINN}. This methodology is often referred to as Physics-Informed Machine Learning \citep{pino, PIDeepOnet, Physics_informed_machine_learning}. We focus on the data-driven perspective; we consider the scenario where sufficient training data is available, and physics constraints serve primarily as a supplementary aid rather than a necessity. There are several challenges that remain unsolved; some are in general for physics-informed machine learning, e.g. how to balance between different losses \citep{pinn_hard1, pinn_hard2, pinn_hard3}. It is worth exploring the potential to enforce certain physics constraints directly through network design, such as Hamiltonian dynamics, conservation laws, symplectic structures, and divergence-free fields. Compared to physics-informed losses, embedding known physics directly into network designs not only guarantees adherence to physics priors but also constrains the model to a smaller search space as shown in Fig. \ref{fig:physical_prior_learning}. Compared to physics-informed losses, this approach not only ensures strict adherence to physical principles but also enhances generalization, improves data efficiency, and leads to interpretable results. Moreover, it can also improve the learning of other coupled physics; a more detailed discussion and an example are given in Sec. \ref{sec:physics_by_design} below.

\begin{tcolorbox}[colback=Apricot, before skip=0.2cm, after skip=0.2cm, boxsep=0.0cm, middle=0.1cm, top=0.1cm, bottom=0.1cm]
\textit{\textbf{Sub-position.} We should design neural operators that enforce physics constraints by design whenever possible. This approach narrows the search space for learning, leading to enhanced interpretability and improved learning of coupled physics.}
\end{tcolorbox}

\subsection{Physics Constraints by Design} \label{sec:physics_by_design}
\begin{figure}[h!]
    \centering
    \begin{minipage}{0.43\textwidth}
        \centering
        \includegraphics[width=0.85\linewidth]{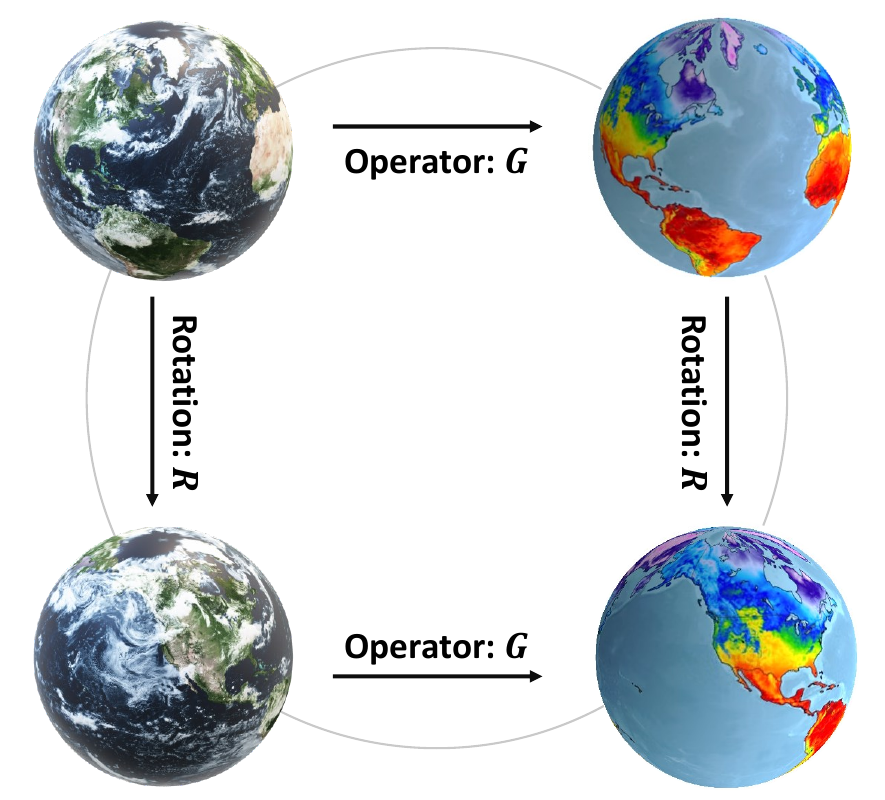}
        \caption{In an operator learning task mapping atmospheric properties (e.g., pressure fields) to the temperature across the globe, rotation symmetries are desired. When you rotate the input atmospheric properties around the globe, the predicted temperature distribution should rotate accordingly.}
        \label{fig:earth_predict_rot}
    \end{minipage}
    \hfill
    \begin{minipage}{0.55\textwidth}
        \centering
        \includegraphics[width=0.99\linewidth]{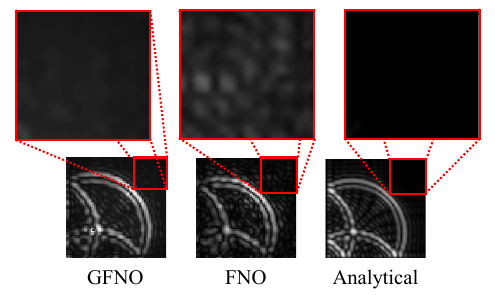}
        \caption{Comparison of the analytical spatial dependencies with the learned spatial dependencies from Group-FNO (GFNO) and FNO. Incorporating known rotation symmetries into FNO enables it to better capture the spatial dependencies.}
        \label{fig:gfno}
    \end{minipage}\vspace{-18pt}
\end{figure}

A notable example of a physics constraint that can be embedded by design is symmetry, which is closely tied to momentum conservation \citep{INO}. Symmetry implies that when the input undergoes a transformation, the output is transformed in a consistent and predictable manner. This property can be mathematically expressed as equivariance:
\begin{equation}
    G\left(\mathscr{L}_h a\right)=\mathscr{L}_h^{\prime} G(a),
\end{equation}
where $G$ is the operator of interest and $\mathscr{L}_h$ and $\mathscr{L}_h^{\prime}$ are the representations of a group element $h$ on the input and output function spaces, respectively. A simple example is rotation equivariance as shown in Fig. \ref{fig:earth_predict_rot}; it should be noted that many symmetry groups are more intricate than rotation symmetry. 

Several studies have explored equivariant network designs for data-driven dynamics. \citet{symmetry_pde_CNN} investigates various symmetries in PDE modeling and equivariance groups like the rotation group are preserved through group convolutions \citep{GroupConv}. Building on this, \citet{helwig2023GFNO} extends group convolution to the FNO architecture, while \citet{TMLR_CTFNO} introduces coordinate transforms to map other symmetries into translational symmetries for convolution-based networks. 
Equivariant designs hold significant potential, as they can greatly enhance the ability to learn other physics, particularly in cases where physics are coupled. Additionally, they can save valuable network capacity from learning symmetries to capturing other critical aspects of physics. For example, the wave equation discussed in Sec. \ref{sec:wave} exhibits rotation symmetries, which are also reflected in the spatial dependencies. We illustrate this by training a Group-FNO \citep{helwig2023GFNO} that is equivariant to the $C_4$ rotation group. The results are presented in Fig. \ref{fig:gfno}; clearly, Group-FNO, with embedded rotation symmetry, demonstrates superior learning of spatial dependencies. It effectively eliminates the noise present in the FNO model outside the arc, revealing clearer and more distinct wave patterns.

Despite their great use and potential in learning data-driven dynamics, the design of equivariant networks for neural operators remains an under-explored area. \textbf{Most existing approaches rely on grid-based representations, making them unsuitable for functional space models that operate on function bases, such as DeepONet.} Alternatively, group averaging or canonicalization provides a model-agnostic approach to achieving equivariance by shifting the burden of equivariance from the model to the data. However, canonicalization may suffer from discontinuity issues that make the learning tasks more difficult for the neural operator \citep{Impossibility_of_cts_Canonicalization}.

\begin{tcolorbox}[colback=Apricot, before skip=0.2cm, after skip=0.2cm, boxsep=0.0cm, middle=0.1cm, top=0.1cm, bottom=0.1cm]
\textit{\textbf{Sub-position.} As known symmetries can improve the learning of underlying physics and enhance the interpretability, future research should focus on developing more advanced equivariant neural operators that support more general symmetry groups and are generally applicable to a broad range of neural operators.}
\end{tcolorbox}

\section{Conclusion and Broader Impacts}
In summary, we believe that neural operators have the potential to uncover hidden physics in learning; however, there is an urgent need to develop explanation methods to assess whether their decision-making process aligns with the underlying physical principles. We classify neural operators into two categories: spatial domain models and functional domain models. In our view, the choice of learning domain or functional bases is critical for capturing specific physics. Building on this classification, we advocate for the design of dual-space, multi-scale models, emphasizing that the focus on resolution invariance or zero-shot super resolution should be reevaluated. Finally, we highlight the importance of incorporating known physics priors into these models to enhance their ability to learn and represent other coupled underlying physical properties effectively for improved interpretability.

This work prompts explainability and interpretability in neural operators as an essential step toward their responsible deployment in high-impact fields such as climate modeling and computational fluid dynamics. Any positive or negative societal impact associated with those applications and domains may also apply to our methods.

\newpage
\bibliography{operator}
\bibliographystyle{plainnat}
\newpage
\appendix
\section{Attention Mechanism as Kernel Integrals} \label{append: attention_no}

Inspired by the remarkable performance of attention mechanisms in natural language processing and computer vision,~\citep{Galerkin_transformer} extends attention mechanisms to the domain of operator learning. This pioneering work laid the foundation for numerous subsequent studies, which have further advanced the use of attention mechanisms in this field~\citep{nonlocal_attention_operator, ONO_emph_resolution_invariance3, transolver}.
The attention mechanism, without considering the softmax function and normalizing constant for simplicity, can be understood as a form of kernel integral:
\begin{equation}
    (\mathcal{K} v)(x)=\int_{\Omega} \left[q(x) \otimes k(y) \right] v(y) \mathrm{d} y, 
\end{equation}
where $k(x, y)=q(x) \otimes k(y)$ is the learned kernel function. 

To be more concrete, we provide a simple realization of such integral layers. Given an $d_{in}$ dimensional input function on a discretization of size $N$, denote by $a_N \in \mathbb{R}^{N \times d_{in}}$. The query $q_N\in \mathbb{R}^{N \times d_{h}}$, key $k_N \in \mathbb{R}^{N \times d_{h}}$, and value $v_N\mathbb{R}^{N \times d_{h}}$, where $d_h$ is the latent dimension, are obtained by linear transformations using trainable weight matrices $W_k, W_q, W_v  \in \mathbb{R}^{d_c \times d_{h}} $:

\begin{equation}
    q_N = a_N W_q, \quad k_N = a_N W_k, \quad v_N = a_N W_v.
\end{equation}

The kernel integral can now be discretized as a summation (numerical integration):

\begin{equation}
(\mathcal{K}v)_N= \sum_{j=1}^N \left[q_N \cdot k_N\right] v_N.
\end{equation}
Note that there are other interpretations and much more advanced realizations of the attention mechanism for operator learning; the readers are referred to those papers~\citep{nonlocal_attention_operator, ONO_emph_resolution_invariance3, transolver, Galerkin_transformer, PIT_emph_resolution_invariance2}.
\section{Spatial Dependencies and the Effective Receptive Field} \label{append:spatial_dependencies_and_wave_analytical}

\textbf{Spatial Dependencies. } Spatial dependencies in PDEs refer to the relationships and interactions between values of the solution at different spatial locations within the domain. These dependencies are governed by the structure of the PDE, the boundary conditions, and the initial conditions if applicable.\cite{Lax2006, PDE_evans10, bucur2016nonlocal} They play a critical role in modeling physical, biological, or engineered systems where phenomena like diffusion, wave propagation, and flow dynamics occur.~\cite{Turing1952,ConstantinFoias1985}. For example, in the diffusion equation ($\left(\nabla^2 u=\frac{\partial u}{\partial t}\right)$), the value of $u$ at a given point is influenced by the surrounding spatial points due to the Laplacian operator $\left(\nabla^2\right)$. In the Navier-Stokes equations, the velocity field at each point in space depends on the surrounding pressure gradients, viscous forces, and external forces, establishing spatial coupling.

Spatial dependencies is also closely related in many deep learning frameworks. For instance, in CNNs, spatial dependencies are captured through local receptive fields that progressively expand via convolutional layers, albeit with limitations in capturing long-range dependencies. Similarly, GNNs can be viewed as more flexible CNNs without the need of regular grid data. Attention mechanisms, such as those in transformers, overcome these limitations by directly modeling global spatial relationships through weighted interactions between all input points. 

For convolution-based networks (e.g., U-Net and FNO), we can study the Effective Receptive Field (ERF)~\citep{Effective_receptive_field}, which determines how a neural network perceives spatial dependencies in the input. Unlike the theoretical receptive field, which represents the entire region of the input contributing to an output unit based on the network architecture, the ERF quantifies the practical influence of each input pixel on the output. Empirically, the ERF tends to occupy only a fraction of the theoretical receptive field due to the network's hierarchical structure and weight distribution. For example, the theoretical receptive field in FNO in global even with only one layers, but for a specific position in the output, the network may not have to pay attention to all the input locations. 

For an output unit, ERF is defined as the pixels in the input that contain a non-negligible impact on that output unit~\citep{Effective_receptive_field}. For simplicity, we assume the one-dimensional (a single channel) input functions $a(x): D \mapsto \mathbb{R}$ and output functions $u(x'): D \mapsto \mathbb{R}$ of the same domain $D$. We adopt the notion of ERF for operators $G: a \mapsto u$ and define the ERF function $\operatorname{erf}: D \times D \mapsto \mathbb{R}$ as the functional derivative: $\operatorname{erf}(x', x) = \frac{\partial u}{\partial a}(x', x)$. For a specific output location $x_o \in D$, we have $\operatorname{erf}_{x_0}(x) = \frac{\partial u}{\partial a}(x', x)|_{x' = x_0}$, which quantifies the sensitivity of the output at a predefined location $x_o$ with respect to perturbations in the input at location $x$ or which locations in the input affect the output prediction the most, thereby capturing the functional dependency structure in neural operators. For grid-based neural operators (e.g. CNO and FNO), given an input function $a(x)$, the ERF function associated with the neural operator can be computed via Autograd location by location.

\section{Experimental Details and Additional Results} \label{append: exp}
\subsection{Wave Equation}\label{append: Wave}
The wave equation from~\citet{convolutional_NO} is given by
\begin{equation}
\begin{aligned}
    u_{tt} - c^2 \Delta u &= 0, && \text{in } D \times (0, T), 
    \\
    u_0(x, y) &= f(x, y), && \text{for } (x, y) \in D.
\end{aligned}
\end{equation}
with a constant propagation speed $c=0.1$ and the domain $D = (0,1) \times (0,1)$. The underlying operator $\mathcal{G}: f \mapsto u(., T)$ maps the initial condition $f$ into the solution at the final time. The initial conditions are given by
\begin{equation} \label{eq: initial_wave}
    f(x, y)=\frac{\pi}{K^2} \sum_{i, j=1}^K a_{i j} \cdot\left(i^2+j^2\right)^{-1} \sin (\pi i x) \sin (\pi j y), \quad \forall(x, y) \in D
\end{equation}
with $K = 24$ and $a_{i j}$ to be i.i.d. uniformly distributed in $[-1,1]$. The exact solution to this wave equation can be computed analytically at any time $t>0$, which is given by
\begin{equation} \label{eq: solution_wave}
    u(x, y, t)=\frac{\pi}{K^2} \sum_{i, j}^K a_{i j} \cdot\left(i^2+j^2\right)^{-1} \sin (\pi i x) \sin (\pi j y) \cos \left(c \pi t \sqrt{i^2+j^2}\right), \quad \forall(x, y) \in D.
\end{equation}
This represents a multiscale standing wave with periodic pulsations in time. 

We are interested in learning the operator $G_t$ mapping the initial condition $u_0$ to the solution $u(\cdot, t)$ at time $t$. In other words, the neural operator is not trained in an auto-regressive manner. The generated dataset contains $1,600$ training trajectories and $400$ testing trajectories. 

We train $5$ different models to learn $G_5$ in Sec.~\ref{sec:wave}:
\begin{itemize}
    \item CNO $3 \times 3$: The original CNO from~\citet{convolutional_NO}, which employs $3 \times 3$ convolution kernels. Slight modifications are made to adapt to this dataset.
    \item CNO $5 \times 5$: The original CNO from~\citet{convolutional_NO} with all $3 \times 3$ convolution kernels replaced by $5 \times 5$ convolution kernels.  Slight modifications are made to adapt to this dataset.
    \item FNO: The original FNO from~\citet{FNO}.  Slight modifications are made to adapt to this dataset.
    \item DeepONet: The modified DeepONet from~\citet{convolutional_NO}. Slight modifications are made to adapt to this dataset.
    \item Galerkin Transformer (GT): The original Galerkin Transformer from~\citet{Galerkin_transformer}. Slight modifications are made to adapt to this dataset.
    \item Transform Once (T1): The original Transform Once from~\citet{T1_NO}. Slight modifications are made to adapt to this dataset.
\end{itemize}
\textbf{Quantitative Results.} The quantitative results are presented in Table~\ref{tab:res_wave}.

\begin{table*}[h!]
    \centering
    \small
    \caption{Results on learning the operator $G_5$ for the wave equation.}
    \begin{tabular}{lcc}
    \Xhline{2\arrayrulewidth}
    \rowcolor{mygray}\multicolumn{1}{c|}{Architecture} & $\#$Parameters (M) & Test $\ell_2$ Error ($\%$)\\ \hline
    CNO $3\times 3$  & 5.84    & 2.28         \\
    CNO $5\times 5$  & 16.19     & 1.63         \\
    FNO      & 1.64     & 2.77      \\
    DeepONet      & 3.21     & 11.82      \\
    GT      & 12.79     & 3.00     \\
    T1      & 1.92     & 5.55 \\
    \Xhline{2\arrayrulewidth}
    \end{tabular}
    \label{tab:res_wave}
    \vspace{-5pt}
\end{table*}

\begin{figure}[h]
    \centering
    \includegraphics[width=0.55\linewidth]{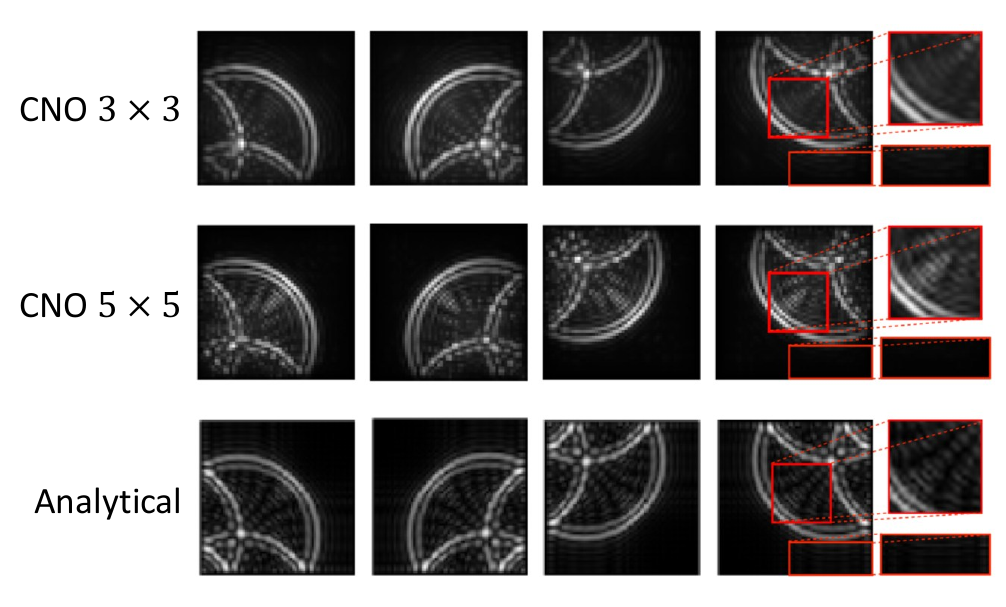}
    \caption{The learned wave patterns of CNO of different local kernel sizes $3\times 3$ and $5 \times5$. While CNO with $5 \times 5$ convolution kernels achieves a better $\ell_2$ error, CNO with $5 \times 5$ convolution kernels does not learn the wave patterns better than that with $3 \times 3$ convolution kernels. } \label{fig: CNO_wave_3vs5}
\end{figure}

\textbf{Additional Qualitative Results.} We provide an additional example showing that standard evaluation metrics, like $\ell_p$ loss, do not necessarily reflect whether the learned neural operators adhere to the underlying dynamics. As shown in Fig. \ref{fig: CNO_wave_3vs5}, while CNO with $5 \times 5$ convolution kernels achieves a better $\ell_2$ error than that with $3 \times 3$ convolution kernels ($1.63\%$ compared to $2.46\%$), CNO with $5 \times 5$ convolution kernels is noticeably over-prone to stronger wave patterns, as the strong waves (e.g. in the highlighted square box) appear excessively bright and the weak wave patterns (e.g. in the highlighted rectangular box) are missing.

\subsubsection{Analytical ERF Function for the Wave Equation} \label{append:analytical_ERF}
For the wave equation given in Appendix~\ref{append: Wave}. We can analytically calculate the associated ERF function through perturbation theory. We are interested in the operator $G_t: u_0(x',y') \mapsto u(x,y,t)$ mapping the initial condition to the solution at time $t$. Consider a small perturbation $\varepsilon_{u_0} \cdot \delta(x - x', y - y')$, we are interested in how the solution changes when the initial condition $u_{0}(x', y')$ is perturbed by this term. Here $\varepsilon$ denotes an infinitesimally small value and $\delta(x - x', y - y')$ is the Dirac delta function defined as 
\begin{equation}
\delta\left(x-x', y-y'\right)= \begin{cases}0 & \text { if }(x, y) \neq\left(x', y'\right) \\ \infty & \text { if }(x, y)=\left(x', y'\right)\end{cases}
\end{equation}
such that
\begin{equation}
\int_{\mathbb{R}^2} \delta\left(x-x', y-y'\right) d x d y=1,
\end{equation}
which is equivalent to
\begin{equation}
\int_{0}^{1}\!\!\int_{0}^{1}
f(x,y)\,\delta(x - x',\,y - y')
\,dx\,dy
\;=\;
f(x',\,y').
\label{eq:delta_definition}
\end{equation}

Specifically, we seek to determine the perturbation on the output, $\varepsilon_{u}$, given the point-wise input perturbation $\varepsilon_{u_0} \cdot \delta(x - x', y - y')$. To proceed, we decompose the Dirac delta function $\delta(x - x', y - y')$ using a basis of sine functions, since $\delta(x - x', y - y') \in L^1$:
\begin{equation}
\delta(x - x', y - y') =\lim_{K\rightarrow \infty} \sum_{i,j=1}^{K} c_{ij} \sin(\pi i x) \sin(\pi j y),
\label{eq:delta_decomposition}
\end{equation}
where $c_{ij}$ are coefficients to be determined. To find $c_{ij}$, we multiply both sides of Eq. \eqref{eq:delta_decomposition} by $\sin(\pi i' x) \sin(\pi j' y)$ and integrate over the domain $[0, 1] \times [0, 1]$:

\begin{equation}
\begin{aligned}
\int_{0}^{1} &\int_{0}^{1} \delta(x - x', y - y') \sin(\pi i' x) \sin(\pi j' y) \, dx \, dy \\&= \lim_{K\rightarrow \infty} \sum_{i,j=1}^{K} c_{ij} \int_{0}^{1} \int_{0}^{1} \sin(\pi i x) \sin(\pi j y) \sin(\pi i' x) \sin(\pi j' y) \, dx \, dy,
\end{aligned}
\label{eq:integral_equation}
\end{equation}
where $i'$ and $j'$ are specific indices. Using the orthogonality property of sine functions:
\begin{equation}
\int_{0}^{1} \sin(\pi p x) \sin(\pi q x) \, dx = 
\begin{cases}
0 & \text{if } p \neq q, \\
\frac{1}{2} & \text{if } p = q,
\end{cases}
\label{eq:orthogonality}
\end{equation}
we obtain the coefficients $c_{ij}$ by Eq. \eqref{eq:delta_definition} :
\begin{equation}
c_{ij} = 4 \int_{0}^{1} \int_{0}^{1} \delta(x - x', y - y') \sin(\pi i x) \sin(\pi j y) \, dx \, dy = 4 \sin(\pi i x') \sin(\pi j y').
\label{eq:cij}
\end{equation}
Substituting $c_{ij}$ back into the decomposition, we have
\begin{equation}
\varepsilon_{u_{0}} \delta(x - x', y - y') = \varepsilon_{u_{0}} \sum_{i,j=1}^{\infty} \sin(\pi i x') \sin(\pi j y') \sin(\pi i x) \sin(\pi j y).
\label{eq:perturbation_decomposition}
\end{equation}
Recall from Sec.~\ref{append: Wave} that the initial condition of the wave equation is given in Eq. \eqref{eq: initial_wave} as
\begin{equation*}
    f(x, y)=\frac{\pi}{K^2} \sum_{i, j=1}^K a_{i j} \cdot\left(i^2+j^2\right)^{-1} \sin (\pi i x) \sin (\pi j y), \quad \forall(x, y) \in D,
\end{equation*}
and the solution of the wave equation is given in Eq. \eqref{eq: solution_wave} as
\begin{equation*} 
    u(x, y, t)=\frac{\pi}{K^2} \sum_{i, j}^K a_{i j} \cdot\left(i^2+j^2\right)^{-1} \sin (\pi i x) \sin (\pi j y) \cos \left(c \pi t \sqrt{i^2+j^2}\right), \quad \forall(x, y) \in D.
\end{equation*}

Therefore, combining Eq. \eqref{eq: initial_wave}, Eq. \eqref{eq: solution_wave}, and Eq. \eqref{eq:perturbation_decomposition} yields
\begin{equation}
\varepsilon_u = \varepsilon_{u_{0}}\lim_{K\rightarrow \infty} \sum_{i,j=1}^{K} \sin(\pi i x') \sin(\pi j y') \sin(\pi i x) \sin(\pi j y) \cos(c \pi t \sqrt{i^{2} + j^{2}}).
\label{eq:delta_u}
\end{equation}
Finally, by perturbation theory,
\begin{equation}
\frac{\partial u}{\partial u_{0}}(x',y', x, y, t) = \frac{\varepsilon_{u}}{\varepsilon_{u_{0}}}= \lim_{K\rightarrow \infty}\sum_{i,j=1}^{K} \sin(\pi i x') \sin(\pi j y') \sin(\pi i x) \sin(\pi j y) \cos(c \pi t \sqrt{i^{2} + j^{2}}).
\label{eq:sensitivity}
\end{equation}

\subsection{Navier-Stokes Equation} \label{append: NS_equation}
The 2D Navier-Stokes equation for a viscous, incompressible fluid in vorticity form from~\citet{FNO} is given by
\begin{equation}
\begin{aligned}
\partial_t w(x, t)+u(x, t) \cdot \nabla w(x, t) & =\nu \Delta w(x, t)+f(x), & & x \in(0,1)^2, t \in(0, T] \\
\nabla \cdot u(x, t) & =0, & & x \in(0,1)^2, t \in[0, T] \\
w(x, 0) & =w_0(x), & & x \in(0,1)^2
\end{aligned}
\end{equation}
where $u \in C\left([0, T] ; H_{\mathrm{per}}^r\left((0,1)^2 ; \mathbb{R}^2\right)\right)$ for any $r>0$ is the velocity field, $w=\nabla \times u$ is the vorticity, $w_0 \in L_{\text {per }}^2\left((0,1)^2 ; \mathbb{R}\right)$ is the initial vorticity, $\nu = 1 \mathrm{e}-3$ is the viscosity coefficient, and $f \in$ $L_{\text {per }}^2\left((0,1)^2 ; \mathbb{R}\right)$ is the forcing function. We are interested in learning the operator that maps the vorticity at the initial time step $t =0$ to the vorticity at a later time $t$ (one-to-one mapping instead of the auto-regressive setting as in~\citet{FNO}). 

Following~\citet{FNO}, the initial condition $w_0(x)$ is generated according to $w_0 \sim \mu$ where $\mu=\mathcal{N}\left(0,7^{3 / 2}(-\Delta+49 I)^{-2.5}\right)$ with periodic boundary conditions. The forcing is kept fixed $f(x)=0.1\left(\sin \left(2 \pi\left(x_1+x_2\right)\right)+\right.$ $\left.\cos \left(2 \pi\left(x_1+x_2\right)\right)\right)$. The equation is solved using pseudospectral method on a $256 \times 256$ grid and then downsample to $64 \times 64$.  The generated dataset contains $1,000$ training samples and $200$ testing samples.

\begin{figure*}[h]
    \centering
    \includegraphics[width=0.9\linewidth]{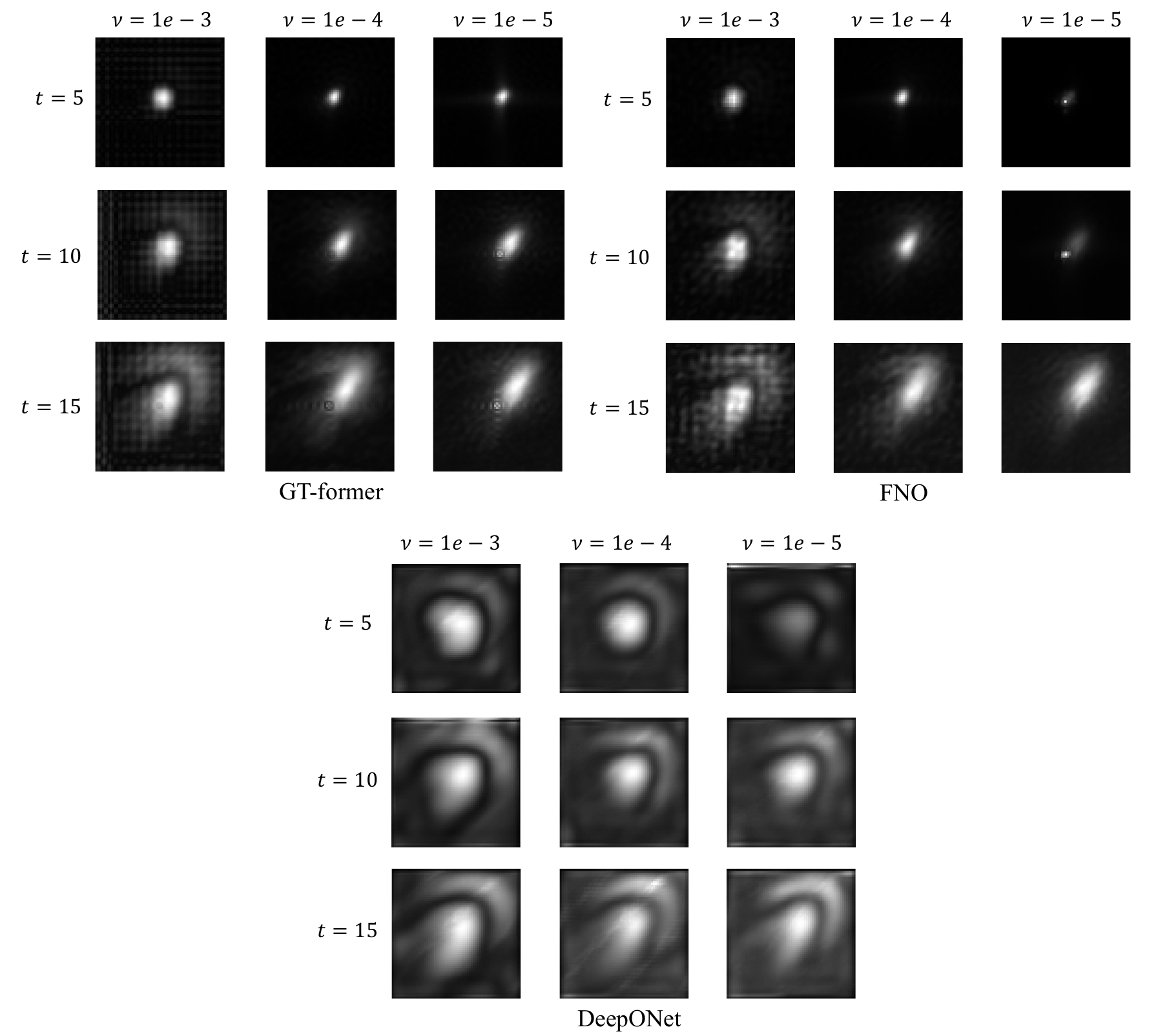}
    \caption{Learned spatial dependencies on the incompressible Navier-Stokes equation reveal insightful patterns. The learned spatial dependencies for the spatial domain model GT-former align with expectations: as the vorticity constant $\nu$ decreases, local patterns become more significant, whereas global patterns gain importance with prolonged time scales. The functional domain model DeepONet does not reveal any meaningful patterns. The hybrid model FNO also learns physically plausible spatial dependencies.} \label{fig:NS_additional}
    \vspace{-12pt}
\end{figure*}

\textbf{Additional Qualitative Results.} We provide additional ERF visualizations for the Navier-Stokes equation using GT-former, FNO, and DeepONet in Fig.~\ref{fig:NS_additional}. Clearly, we can observe that the spatial domain model, GT-former, better captures the spatial dependencies, while the functional domain model, DeepONet, does not produce any meaningful patterns. The hybrid model FNO also learns physically plausible spatial dependencies.

\subsection{Darcy Flow Equation}\label{append:darcy_data}
We consider the steady-state of the 2D Darcy Flow equation from~\citet{FNO} given by:
\begin{equation}
\begin{aligned}
-\nabla \cdot(a(x) \nabla u(x)) & =f(x) & & x \in(0,1)^2 \\
u(x) & =0 & & x \in \partial(0,1)^2
\end{aligned}
\end{equation}
where $a \in L^{\infty}\left((0,1)^2 ; \mathbb{R}_{+}\right)$ is the diffusion coefficient and $f \in$ $L^2\left((0,1)^2 ; \mathbb{R}\right)$ is the forcing function that is kept fixed $f(x)=1$. We are interested in learning the operator mapping the diffusion coefficient $a(x)$ to the solution $u(x)$. The dataset is provided by~\citet{FNO}. The input diffusion coefficient field $a(x, y)$ is generated by the Gaussian random field with a piecewise function, namely $a(x, y)=\psi(\mu)$, where $\mu$ is a distribution defined by $\mu=\mathcal{N}\left(0,(-\Delta+9 I)^{-2}\right)$. The mapping $\psi: \mathbb{R} \rightarrow \mathbb{R}$ takes the value $12$ on the positive and $3$ on the negative, and the push-forward is defined point-wise. Solutions are obtained using a second-order finite difference scheme on a $421 \times 421$ grid and then downsampled to $85 \times 85$. The dataset used, including training/testing split, is exactly the same as in~\citet{FNO}.

\begin{figure}[t]
    \centering
    \includegraphics[width=0.75\linewidth]{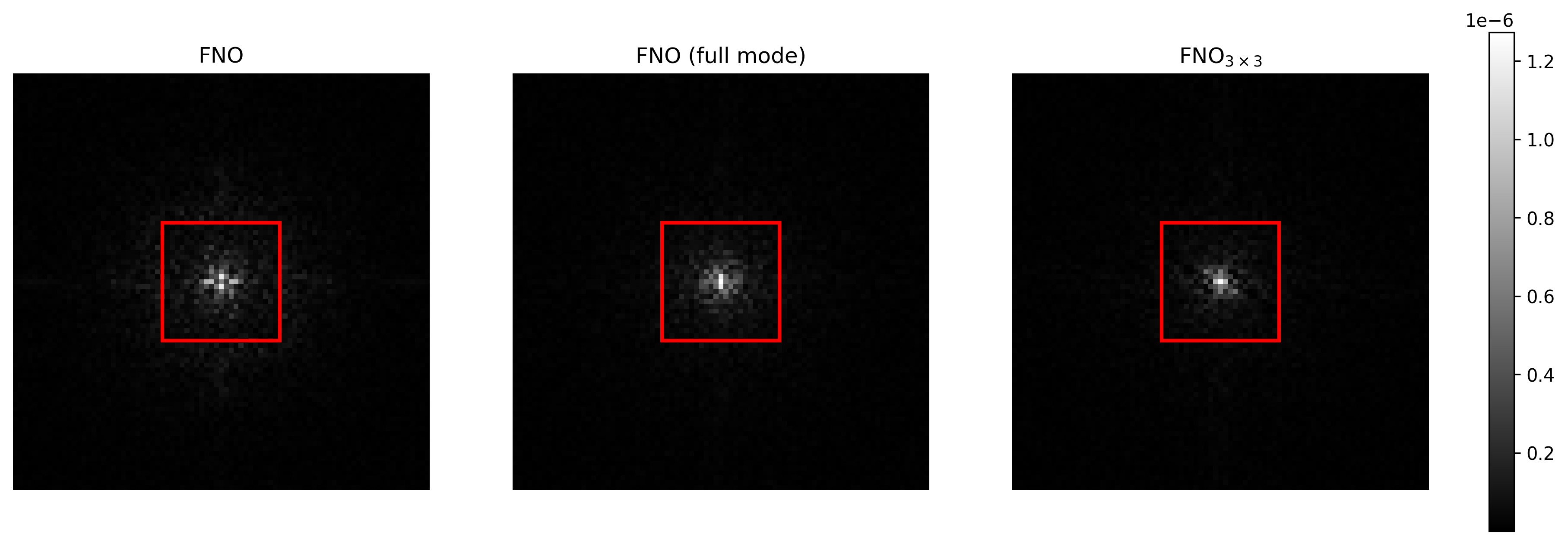}
    \caption{Spectral plot of the error distribution for FNO, FNO$_{\text{full}}$, and FNO$_{3\times 3}$. Red boxes indicate the truncation applied by the base FNO model. Clearly, local convolution kernels reduce errors not only in the high-frequency modes but also in the low-frequency modes.} \label{fig:spectral_multiscale}
\end{figure}

In addition, we provide the spectral plot of the error distribution in Fig~\ref{fig:spectral_multiscale} to show that the inclusion of local convolution kernels also reduces errors from low-frequency modes. This straightforward and simple modification demonstrates significant improvement, highlighting the potential for further exploration and refinement.

\subsection{Helmholtz Equation}

We consider the 2D Helmholtz equation from \citet{helm1} given by 
\begin{equation}
\begin{aligned}
\left(-\Delta-\frac{\omega^2}{c^2(x)}\right) u & =0 & & \text { in } \Omega, \\
\frac{\partial u}{\partial n} & =0 & & \text { on } \partial \Omega_1, \partial \Omega_2, \partial \Omega_4, \\
\frac{\partial u}{\partial n} & =u_N & & \text { on } \partial \Omega_3 ,
\end{aligned}
\end{equation}
where $\Omega = [0,1]^2$ and $\{\Omega_i\}_{i = 1}^4$ are the four edges of the square domain. $\omega$ is set to $10^3$, $c: \Omega \rightarrow \mathbb{R}$ is the wave-speed field, and $u: \Omega \rightarrow \mathbb{R}$ is the excitation field that solves the equations. The Neumann boundary condition imposed on $\partial \Omega$ is non-zero only on the top edge $\Omega_3$. Throughout the experiments presented in this work, $u_N$ is fixed at $1_{\{0.35 \leq x \leq 0.65\}}$ following \citet{helm1}. The wave-speed field $c(x)$ is assumed to be
\begin{equation}
c(x)=20+\tanh (\tilde{c}(x)),
\end{equation}
where $\tilde{c}$ is a centered Gaussian
\begin{equation}
\tilde{c} \sim \mathbb{N}(0, \mathrm{C}) \quad \text { and } \quad \mathrm{C}=\left(-\Delta+\tau^2\right)^{-d}.
\end{equation}

Here, $-\Delta$ denotes the Laplacian operator on $D_u$ under homogeneous Neumann boundary conditions, restricted to functions with zero spatial mean. The parameters are set to $d = 2$ and $\tau = 3$. Our objective is to learn the mapping from the wave-speed field $c$ to the solution $u$. The dataset used in this work, provided by \citet{helm1}, is generated by solving the Helmholtz equation using the finite element method on a $100 \times 100$ grid. We use the first 800 samples as training data and the next 200 samples for testing. The data is normalized to the $[0, 1]$ range using statistics from the training set.

\subsection{Allen-Cahn Equation}
We consider the Allen-Cahn equation from~\citet{convolutional_NO} given by 
\begin{equation}
\begin{aligned}
u_t(x, t)& =\Delta u(x, t)-\varepsilon^2 u(x, t)\left(u(x, t)^2-1\right),~ t>0 \\
 u(x, 0)& =f(x),
\end{aligned}
\end{equation}
with a reaction rate of $\varepsilon=220$. The goal is to learn the underlying operator $\mathcal{G}: f(\cdot) \mapsto u(\cdot, T)$, mapping initial conditions $f$ to the solution $u$ at a final time $T=0.0002$. The initial conditions are given by
\begin{equation} 
    f(x, y)=\frac{\pi}{K^2} \sum_{i, j=1}^K a_{i j} \cdot\left(i^2+j^2\right)^{-1} \sin (\pi i x) \sin (\pi j y), \quad \forall(x, y) \in D
\end{equation}
with $K = 24$ and $a_{i j}$ to be i.i.d. uniformly distributed in $[-1,1]$. Analytical solutions are not available. Both training and test data are generated by using a finite difference scheme on a grid of $64\times 64$ resolution.
\subsection{Experiments Compute Resources}
The experiments are conducted on an NVIDIA RTX A6000 GPU with 48 GB of GDDR6 memory. As this is a position paper, we do not include extensive experiments; each neural operator in the numerical examples can be trained within an hour for a single run. The computational cost of the ERF visualization is minimal.

\subsection{Licenses for Existing Assets}
We list all the licenses for existing assets in Table~\ref{tab:assets-licenses-description}.
\begin{table}[t] 
    \centering
    \caption{Assets, Licenses, and Descriptions}
    \resizebox{\columnwidth}{!}{
    \begin{tabular}{lll}
        \toprule
        \textbf{Asset} & \textbf{License} & \textbf{Description} \\
        \midrule
        FNO~\citep{FNO} & MIT License & Neural Operator Model \\
        CNO~\citep{convolutional_NO} &  MIT License & Neural Operator Model \\
        Galerkin Transformer~\citep{Galerkin_transformer} & MIT License & Neural Operator Model\\
        DeepONet~\citep{DeepOnet}& CC BY-NC-SA 4.0 & Neural Operator Model\\
        T1~\citep{T1_NO}& MIT License & Neural Operator Model\\
        The Wave Equation Dataset~\citep{convolutional_NO} & \textit{N/A} & Dataset\\
        The Navier-Stokes Equation Dataset~\citep{FNO} & MIT License & Dataset\\
        The Darcy Flow Equation Dataset~\citep{FNO} & MIT License & Dataset\\
        The Helmholtz Equation Dataset~\citep{helm1}& CC-0 & Dataset\\
        The Allen-Cahn Equation Dataset~\citep{convolutional_NO} & MIT License & Dataset\\
        \bottomrule
    \end{tabular}
    }
\label{tab:assets-licenses-description}
\end{table}

\end{document}